\documentclass[conference]{IEEEtran}
\IEEEoverridecommandlockouts
\usepackage{cite}
\usepackage{amsmath,amssymb,amsfonts}
\usepackage{algorithmic}
\usepackage{graphicx}
\usepackage{textcomp}
\usepackage{xcolor}
\usepackage[ruled,linesnumbered]{algorithm2e}

\def\BibTeX{{\rm B\kern-.05em{\sc i\kern-.025em b}\kern-.08em
    T\kern-.1667em\lower.7ex\hbox{E}\kern-.125emX}}
\begin{document}

\title{
    Exploring Adversarial Robustness of LiDAR-Camera Fusion Model
    in Autonomous Driving\\

}

\makeatletter
\newcommand{\linebreakand}{%
  \end{@IEEEauthorhalign}
  \hfill\mbox{}\par
  \mbox{}\hfill\begin{@IEEEauthorhalign}
}
\makeatother

\author{\IEEEauthorblockN{Bo Yang}
\IEEEauthorblockA{\textit{College of Electrical Engineering} \\
\textit{Zhejiang University}\\
Hangzhou, Zhejiang \\
yb5@zju.edu.cn}
\and
\IEEEauthorblockN{Xiaoyu Ji}
\IEEEauthorblockA{\textit{College of Electrical Engineering} \\
\textit{Zhejiang University}\\
Hangzhou, Zhejiang \\
xji@zju.edu.cn}
\and
\IEEEauthorblockN{Zizhi Jin}
\IEEEauthorblockA{\textit{College of Electrical Engineering} \\
\textit{Zhejiang University}\\
Hangzhou, Zhejiang \\
zizhi@zju.edu.cn}
\linebreakand
\IEEEauthorblockN{Yushi Cheng}
\IEEEauthorblockA{\textit{College of Electrical Engineering} \\
\textit{Zhejiang University}\\
Hangzhou, Zhejiang \\
yushicheng@zju.edu.cn}
\and
\IEEEauthorblockN{Wenyuan Xu}
\IEEEauthorblockA{\textit{College of Electrical Engineering} \\
\textit{Zhejiang University}\\
Hangzhou, Zhejiang \\
wyxu@zju.edu.cn}

}

\maketitle

\begin{abstract}
    Our study assesses the adversarial robustness of LiDAR-camera 
    fusion models in 3D object detection. We introduce an attack 
    technique that, by simply adding a limited number of physically 
    constrained adversarial points above a car, can make the car 
    undetectable by the fusion model. Experimental results reveal 
    that even without changes to the image data channel—and solely 
    by manipulating the LiDAR data channel—the fusion model 
    can be deceived. This finding raises safety concerns in the 
    field of autonomous driving. Further, we explore how the 
    quantity of adversarial points, the distance between the 
    front-near car and the LiDAR-equipped car, and various angular 
    factors affect the attack success rate. We believe our research 
    can contribute to the understanding of multi-sensor robustness, 
    offering insights and guidance to enhance the safety of 
    autonomous driving.

\end{abstract}


\section{Introduction}
In recent years, automotive companies have been enhancing 
the level of automation and intelligence in their vehicles, 
and the market's acceptance of Autonomous Driving (AD) has 
grown correspondingly. To earn trust in AD, 
ensuring its safety is paramount. A primary component of 
AD is visual perception, which forms the basis for 
dependable decision-making. The AD system employs sensors 
like cameras and LiDAR to gather data about the 
surrounding environment. This data is then processed 
through deep learning models to achieve perceptions 
such as object detection.

Cameras and LiDAR are the predominant sensors in AD systems. 
Both have their own strengths and weaknesses: 
the more affordable cameras can provide high-resolution images 
with rich texture details, but are limited in view and depth information, 
whereas the pricier LiDAR can offer 360° point cloud data 
with rich depth insights, but the data is unordered and sparse.

\begin{figure}[t]
	\centering
	\includegraphics[width=0.48\textwidth]{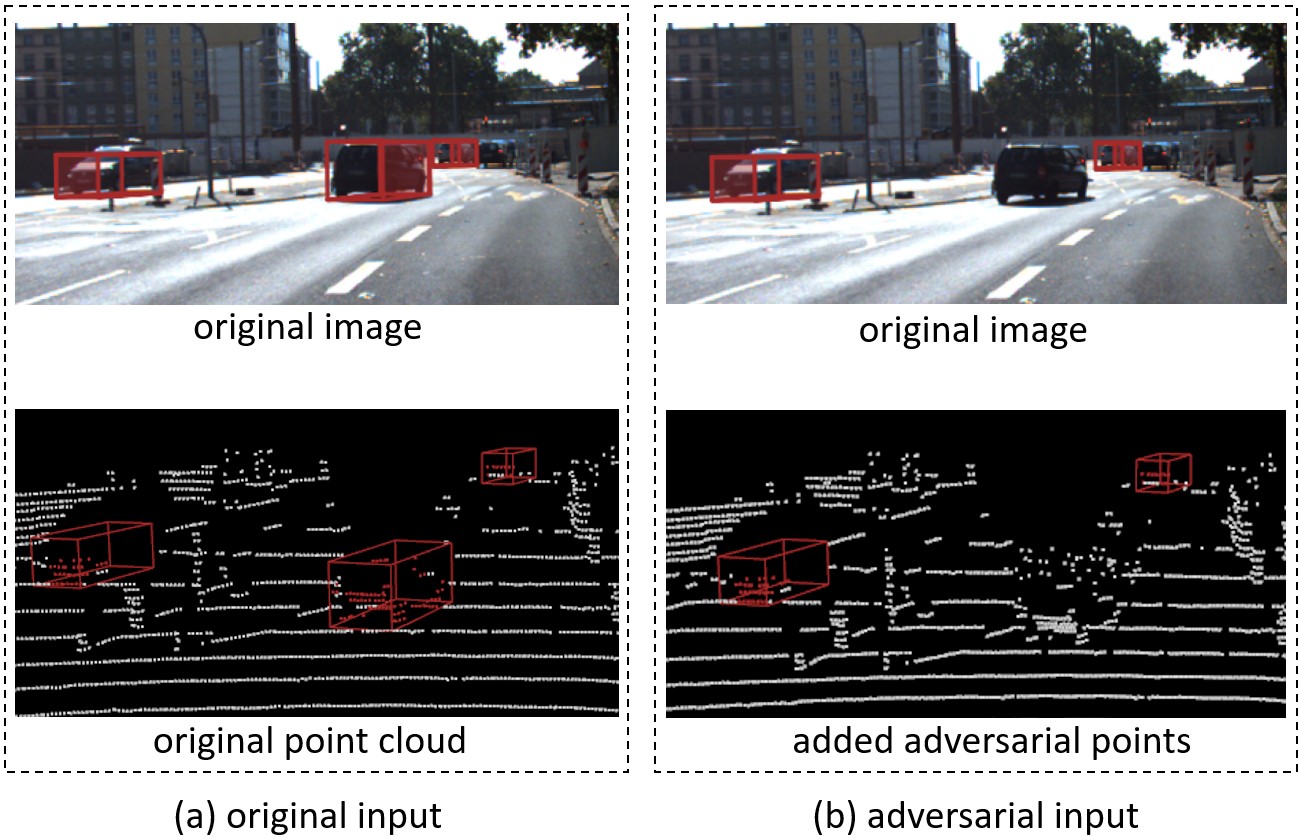}
	\caption{Adversarial points, when subjected to physical constraints, 
    can deceive the LiDAR-camera fusion 3D detection model used in 
    autonomous driving. By adding a limited number of these 
    adversarial points above the central car, it becomes 
    undetectable by the fusion model.}
	\label{intro}
	\vspace{-5pt}
\end{figure}

The AD system typically employs deep neural network-based models 
to process data collected by sensors, facilitating perception 
tasks such as object detection. However, deep neural networks 
have been widely demonstrated to be vulnerable, particularly 
to adversarial samples 
~\cite{szegedy2013intriguing,carlini2017towards,goodfellow2014explaining}.
Initially, some studies focused on the safety of camera-only models in AD. 
Examples include placing patches on stop signs~\cite{eykholt2018robust} 
or adding misleading dirt patches on roads~\cite{274691}. 
The security challenges related to LiDAR-only models in AD 
have also garnered attention. Such investigations can be categorized 
into two approaches: attackers injecting adversarial points into 
victim LiDAR via a physical-world laser 
transmitter~\cite{cao2019adversarial, wang2021adversarial, 
sun2020towards, wang2021adversarial, jin2023pla}, and the creation 
or positioning of adversarial 3D objects
~\cite{tu2020physically, yang2021robust, zhu2021can}.
However, since both cameras and LiDAR have their own strengths 
and weaknesses, a natural idea is to leverage the advantages 
of both to complement each other. As a result, LiDAR-camera fusion models
like Waymo’s One and Baidu’s Apollo, have emerged. 
Increasingly, research is addressing the security of these fusion models.
In the LiDAR-camera fusion model, data from both sensors 
serve as the model's input. This presents two data 
channels—image and points—for potential exploitation by 
attackers. Some research has focused on introducing disturbances 
only to image data to impact the fusion model's object 
detection capabilities~\cite{park2021sensor}, others 
on injecting malicious points solely into LiDAR
~\cite{hallyburton2022security}, and still others on 
strategically positioning objects with adversarial shapes 
and colors to simultaneously deceive both data channels
~\cite{tu2021exploring, abdelfattah2021adversarial, cao2021invisible}.
However, in a real-world setting, it's challenging to disrupt 
the image input of the fusion model in an AD system~\cite{park2021sensor}. 
Moreover, using 3D objects with adversarial features might 
easily draw the attention of potential victims~\cite{cao2021invisible}. 
It's also tough to guarantee effective deception by introducing 
random point clouds with an orthographic distribution into the 
image's field of view~\cite{hallyburton2022security}.

In this work, we aim to explore the security issues associated with the 
LiDAR-camera fusion model when subjected to adversarial points that 
satisfy physical constraints. Specifically, for the adversarial points 
generated at the software level to be injected into the LiDAR via a 
physical-world laser transmitter, the generation process must adhere 
to the following three physical constraints:
(1)Each laser ray can record at most one point.
(2)The vertical angles of adversarial points must align with the laser ray's discrete specific angles.
(3)The horizontal angle of adversarial points must be restricted to a narrow range.
Using adversarial machine learning methods, we generate points 
designed to deceive the fusion model. Specifically, we introduce 
a minimal number of adversarial points above the target vehicle, 
preventing its detection by the fusion model (referred to as the 
"Hiding Attack" or HA). Such a strategy may lead to real-world traffic 
hazards, such as rear-end collisions.

We selected the classic fusion model, MVX-Net~\cite{sindagi2019mvx}, 
to explore its robustness when injected with adversarial points that 
satisfy physical constraints in the realm of autonomous driving. 
Experimental results indicate that the LiDAR-camera fusion model 
remains vulnerable to adversarial points adhering to these physical 
constraints, even without changes to the image data channel. 
Furthermore, we delve into how the number, distance, and angle of injected 
adversarial points affect the performance of the fusion model.

\section{Background}

\subsection{Perception in Autonomous Driving}
Perception is the foundational element of autonomous driving, 
bridging the gap between the vehicle and its environment. 
Its purpose is to mimic the human eye, collecting relevant information 
and providing essential data for subsequent decision-making. 
The overall performance of an autonomous driving system is heavily 
influenced by the effectiveness of its perception system. 
Within this framework, object detection (such as detecting 
pedestrians and cars) serves as a crucial component. 
To achieve this functionality, autonomous driving vehicles 
are equipped with multiple sensors across various modalities, 
along with object detection algorithms that transform sensor data 
into meaningful semantic information.

\subsection{LiDAR-Camera Fusion 3D Object Detection}
In the field of autonomous driving, LiDAR and cameras are the 
two most common sensors utilized to detect objects in the 
surrounding environment. However, the data they collect each has its 
unique strengths and weaknesses. Image data, though rich in texture 
details, is constrained by its field of view and lacks depth information. 
On the other hand, the 360-degree point cloud data generated by 
LiDAR offers valuable depth insights but is disordered and sparse.
By fusing data from both LiDAR and cameras, their individual shortcomings 
can be addressed, leading to a more comprehensive three-dimensional 
environmental representation. Consequently, object detection based on 
this fusion is gaining traction, especially in high-level AD 
vehicles~\cite{apollo}.
Existing LiDAR-camera fusion model structures for object detection 
tasks can be broadly categorized into three types
~\cite{hallyburton2022security}:
Cascaded semantic fusion, which utilizes the output of perception 
from one or multiple sensors to augment the input of other 
single-sensor perceptions~\cite{qi2018frustum, wang2019frustum}.
Integrated semantic fusion, where isolated perception operations 
are conducted for each sensor and their semantic outputs are then 
fused~\cite{apollo}.
Feature-level fusion, which employs Deep Neural Networks (DNNs) 
to simultaneously extract and merge image and point features. 
The resultant combined representation is then processed through a 
DNN for 3D detection~\cite{sindagi2019mvx, huang2020epnet, ku2018joint}.

\begin{figure}[t]
	\centering
	\includegraphics[width=0.45\textwidth]{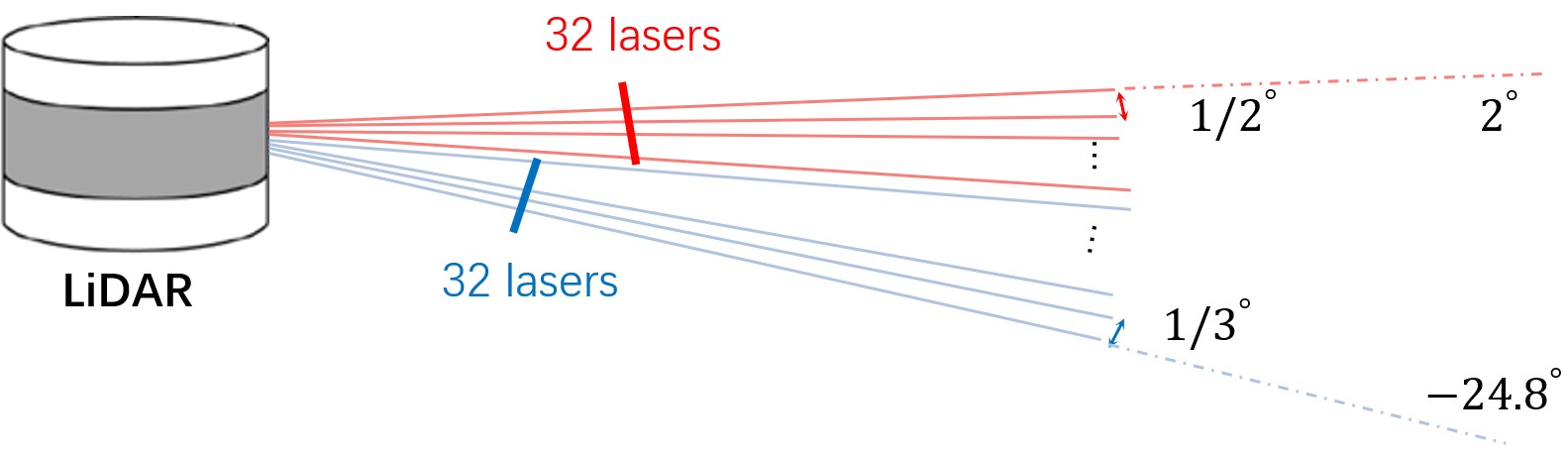}
	\caption{Velodyne HDL-64E has 64 lasers, and their vertical angle range
     is -24.8° to  2°. 32 lasers in lower laser block separated by
      1/2° vertical spacing and 32 lasers in upper laser block separated
       by 1/3° vertical spacing.}
	\label{Velodyne64LiDAR}
	\vspace{-5pt}
\end{figure} 
\begin{figure*}[pt]
    \centering
    \includegraphics[width=0.85\textwidth]{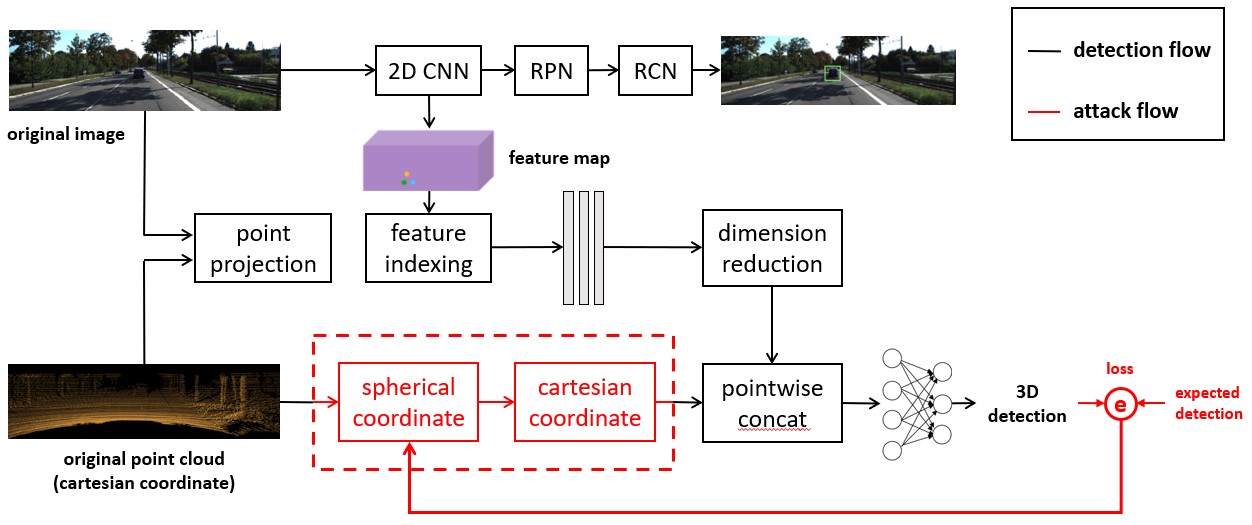}
      \vspace{-5pt}
    \caption{Overview of the attack pipeline.
    The standard data processing flow of MVX-Net is indicated by 
    solid black lines. To ensure the injectability of adversarial 
    points into the victim LiDAR in the real world, these points 
    are transformed based on a specific angle in the spherical 
    coordinate system, using the LiDAR as the reference point. 
    This additional data processing step is denoted by a solid red line. 
    The flow of gradients is shown by dashed red lines. 
    As the entire pipeline is differentiable, gradients can flow 
    from the adversarial loss back to the adversarial points 
    within the spherical coordinate system.
    Ultimately, cars with adversarial points situated above their roofs are 
    rendered undetectable by the fusion model.
}
    \label{attack_overview}
    \vspace{-5pt}
\end{figure*}
\subsection{Adversarial Attack in AD}
Recently, there has been a surge in interest around adversarial attacks 
that exploit the vulnerabilities of machine learning algorithms. 
Researchers have proposed various methods to create adversarial 
examples (or images) that can lead to misclassifications in 2D image 
classification and object detection
~\cite{szegedy2013intriguing,carlini2017towards,goodfellow2014explaining}. 
With the rise of Autonomous Driving (AD), an increasing amount of 
research is focusing on the safety of perception within the AD system. 
In particular, adversarial machine learning targeting LiDAR-based 3D 
object detection is garnering attention
~\cite{cao2019adversarial, tu2020physically, jin2023pla}. 
Some studies have delved deeper into such attacks in real-world settings
~\cite{eykholt2018robust, 274691, cao2019adversarial, 
jin2023pla, yang2021robust, zhu2021can, hallyburton2022security}.
In the AD domain, prior research has successfully executed 
real-world adversarial attacks on camera-based AD perception, 
such as placing patches on stop signs~\cite{eykholt2018robust} 
or deploying deceptive road patches~\cite{274691}. 
Other works have targeted LiDAR-based perceptions, 
either by injecting adversarial points into the target LiDAR via 
laser transmitters~\cite{cao2019adversarial, jin2023pla} 
or by designing and positioning adversarial 3D objects
~\cite{yang2021robust, zhu2021can}. Additionally, 
some research has explored the impact on target detection 
accuracy within LiDAR-camera fusion models, particularly by 
introducing adversarial points into the image cone
~\cite{hallyburton2022security}.

\section{Attack Design}
In this work, we aim to place adversarial points atop a car within 
a 3D scene, ensuring there's no occlusion, with the goal of 
evading detection by the LiDAR-camera fusion model. We only modify 
the LiDAR data input channel of the fusion model. The physical 
constraints for the adversarial points, as determined by the LiDAR 
acquisition data, are maintained by merely adjusting the adversarial 
points' coordinates within the spherical coordinate system 
(see Fig. \ref{attack_overview}). We employ this attack to analyze 
how these adversarial points impact the car detection accuracy of 
an AD's LiDAR-camera 3D detection model. Our focus is on car detection, 
given its significance as one of the most safety-critical tasks 
in an AD's perception system. In the following sections, 
we elaborate on the threat model, physical constraints, victim models, 
attack objective function, and the optimization process.

\subsection{Threat Model}
We assume an attack scenario as follows: The attacker hides in a 
concealed position on the road and uses laser transmitters to 
inject adversarial points into passing vehicles. 
The goal is to deceive the LiDAR-camera fusion model deployed on 
the victim vehicle, rendering the front-near cars undetectable. 
Such a deception could lead to severe traffic incidents, 
including rear-end collisions. Furthermore, the attacker is 
knowledgeable about adversarial machine learning and has access 
to the fusion model's relevant parameters. They can also obtain 
the model's output results (e.g., Apollo is an open source autopilot
framework). Specifically, the adversary 
has the capability to consistently inject over 200 adversarial 
points, with the horizontal angle range of the injected adversarial 
point cloud being less than 10°.

\subsection{Physical Constraint}
The point cloud space is sparse. Every generated point can only 
reside on one of the sparse LiDAR's laser rays, and each laser ray 
can record at most one point. Therefore, to ensure that the LiDARs 
can perceive adversarial points emitted by the adversary's transmitter 
in the real world, we must ensure that the added adversarial points 
align with these laser rays. Specifically, we adhere to the following 
physical constraints when generating adversarial points:

\begin{itemize}
\item \textbf{Each laser ray can record at most one point.} 
Autonomous vehicles equipped with LiDARs often operate under the 
Strongest Return Mode setting~\cite{cao2019adversarial}. 
In this mode, each laser only records the point with the highest 
reflection intensity. Without proper constraints on the 
adversarial points, there might arise a situation where two 
adversarial points appear on the same laser ray, 
violating physical reality.
\item \textbf{The vertical angles of the adversarial points 
must align with the laser ray's discrete specific angles.} 
Common mechanical LiDARs used in AD come in configurations 
like 16-line, 32-line, 64-line, and so on (refer to 
Fig. \ref{Velodyne64LiDAR}). These laser beams are distributed 
at specific vertical angles centered on the LiDAR. As a 
result, it's crucial to ensure that the added adversarial 
points fall within these preset angles.
\item \textbf{The horizontal angle of the adversarial points 
must be restricted to a narrow range.} The attacker's capabilities 
are bound by hardware constraints, such as the laser transmitter. 
As such, it's infeasible to inject adversarial points across the 
full 360° horizontal angle of the target LiDAR simultaneously. 
Currently, attackers can generally introduce adversarial points 
within a 10° horizontal angle distribution.
\end{itemize}

\subsection{Victim Models}
We selected the MVX-Net model~\cite{sindagi2019mvx} for our experiments 
to investigate its detection performance in the presence of adversarial 
points constrained by physical limitations. MVX-Net is a classic 
feature-level fusion model. It employs convolutional filters from a 
pre-trained 2D faster RCNN to compute the image feature map. 
The 3D points are projected onto the image using calibration data, 
and the corresponding image features are appended to these 3D points.

\subsection{Attack Objective Function}
The adversaryis goal is to add adversarial points atop the targeted object, 
making it undetectable. Our strategy involves suppressing 
bounding box proposals related to the targeted objects. 
A proposal near the targeted object is considered relevant 
if it meets two criteria: its Intersection over Union (IoU) is 
above a specified threshold, denoted as \(\epsilon_i\), and 
the confidence level in class prediction for the proposal exceeds a 
certain threshold, denoted as \(\epsilon_s\). 
Considering the complexities of real-world attacks, 
we choose to place adversarial points above the targeted object. 
This approach aims to suppress relevant proposals while avoiding 
possible occlusions or interferences caused by the targeted object itself. 
Therefore, we have devised our attack objective function in this manner:

\begin{equation}
    \begin{split}
        \mathcal{L}_{adv} =\mathop{\Sigma}_{p, c \in { P}} { \rm IoU}(p^{gt},p){\rm log}(c) 
        \label{adversarial_loss}     
    \end{split}
\end{equation}
where \( P = \{(x_i,y_i,z_i,d_i,w_i,h_i,rot_i,c_i) | i \in [1,n]  \} \) 
represents the set of all bounding box proposals; 
\( n \) is the total number of these adversarial point proposals; 
\( p^{gt} \) denotes the actual position of the targeted object, and 
\( p \) and \( c \) are the the relevant bounding box
proposals and their confidences, respectively 
In our approach, $\epsilon_i=0.1$ and $\epsilon_s=0.1$.


\begin{algorithm}[t]
	\caption{Generating adversarial points under physical constraints}
	\label{advprocess}
	\KwIn{clean point cloud : $P$\\
		\quad expected location : $ {\rm Loc}^{exp} $\\
		\quad number of adversarial points: $ n $ \\
        \quad number of initialization times : $ n_{init} $ \\
		\quad number of optimization iteration : $ n_{iter} $}
    \For { $j \in \{1, 2, 3,..., n_{init}$\}} {
        randomly initialize adversarial points $P_{adv}  = \{(R_i^{'},\alpha_i^{'},\omega_i^{'}) \in  {\rm Loc}^{exp}  | i \in [1,n] \}$\\
        $P^{'} \gets P + P_{adv}$ \\  
        \For{ $k \in \{1,2,3,..., n_{iter}$\} }{
            convert $P_{adv}$ to spherical coordinate $P_{advs}$\\
            convert $P_{advs}$ to Cartesian coordinates $P_{adv}$   \\       
            get detection result: ${\rm M} (P_a )$\\
            if $ {\rm Obj}^{exp} $ undetected :  \\
            \quad    break  \\
            calculate $\delta^{'}$ by function \ref{hiding_loss}\\
            update $P_{advs} \gets P_{advs} + \delta^{'}$ \\
            convert $P_{advs}$ to Cartesian coordinates $P_{adv}$   \\
            update $P^{'} \gets P + P_{adv}$ \\   
        }	
        if $ {\rm Obj}^{exp} $ undetected : \\
        \quad   break \\
    }

	\KwOut{adversarial point cloud : $P_{adv}$}
\end{algorithm}

\begin{figure}[t]
	\centering
	\includegraphics[width=0.45\textwidth]{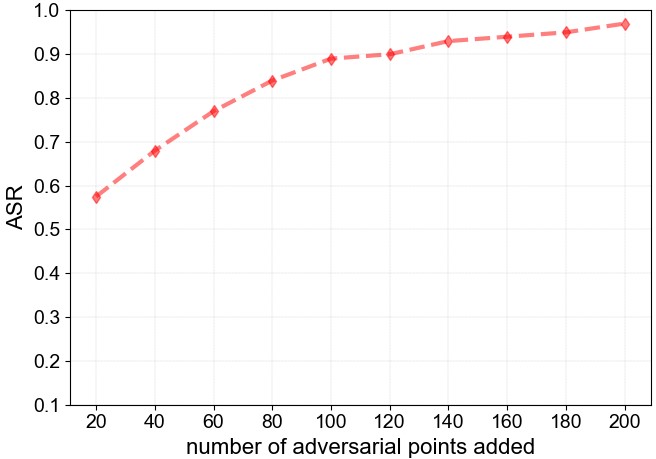}
	\caption{Attack success rate of spoofing fusion model to 
    make front-near cars undetected
    with different number of adversarial points. 
    The ASR roughly increases with the number of added adversarial points.}
	\label{pointsnum}
	\vspace{-5pt}
\end{figure} 

\begin{figure}[t]
	\centering
	\includegraphics[width=0.45\textwidth]{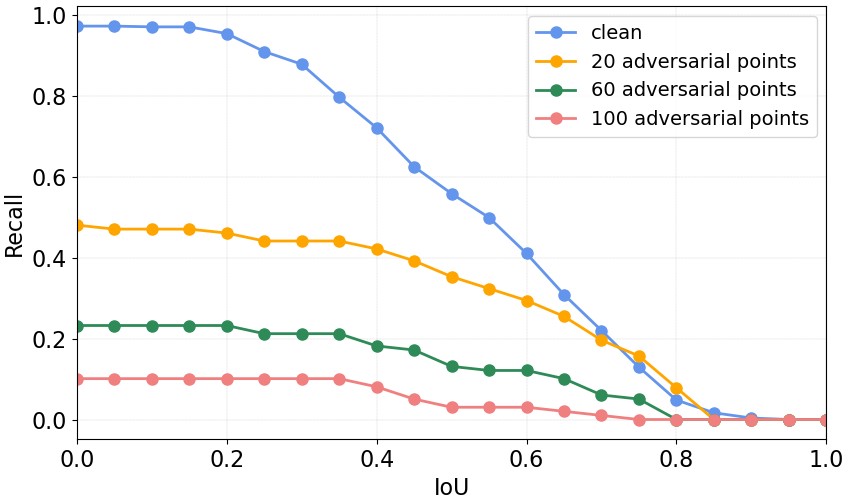}
	\caption{Recall-IoU curves with different IoU thresholds.
    The curve labeled with "clean" represents the performance 
    of the fusion model MVX-Net~\cite{sindagi2019mvx} without interference from 
    adversarial points that comply with physical constraints.}
	\label{recalliou}
	\vspace{-5pt}
\end{figure}

\subsection{Optimization Process}
The process for generating adversarial points, while adhering to 
physical constraints, is outlined in Algorithm \ref{advprocess}. 
We restrict the space for adding adversarial points to a 
3.6 $\times$ 3.6 m $\times$ 1m region above the car we aim to obscure. 
Within this space, we calculate the allowable variations in distance, 
vertical angle, and horizontal angle for the adversarial points 
based on spherical coordinates.

During optimization, to ensure the adversarial points comply with 
physical constraints, we update the distance of these points using 
the gradient information from the function's backpropagation within 
the spherical coordinate system. However, we do not adjust the vertical 
and horizontal angles of the adversarial points based on gradient 
information. This constraint ensures that each laser ray captures at 
most one point.

The initial vertical and horizontal angles of the adversarial points 
in the spherical coordinate system are set during random initialization. 
To mitigate the impact of this randomness, we permit the adversarial 
points to be re-initialized $ n_{init} $ times during the 
optimization process.

\begin{figure}[t]
	\centering
	\includegraphics[width=0.42\textwidth]{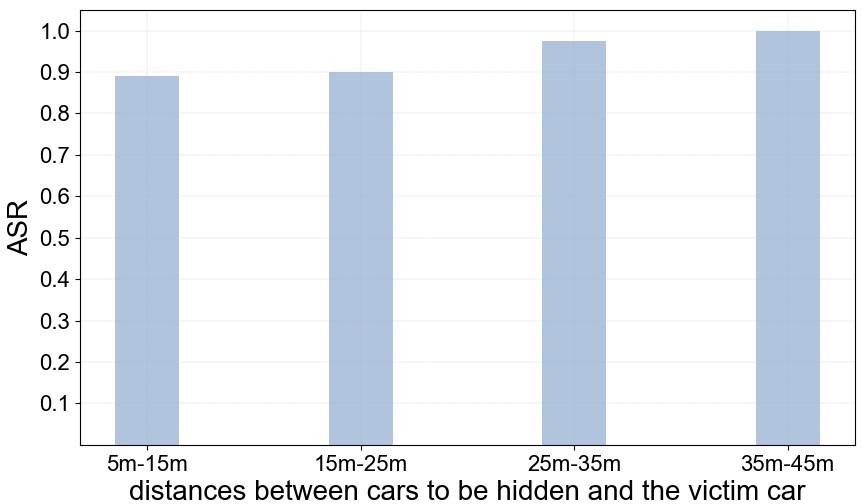}
	\caption{Attack success rate of spoofing fusion model
    with different distances between cars to be hidden and
    the victim car deployed with LiDAR.}
	\label{distance}
	\vspace{-5pt}
\end{figure} 

\section{Evaluation}
We assess the efficacy of adversarial points under physical 
constraints in deceiving the LiDAR-camera fusion model. 
Our focus is on car detection, a safety-critical task in AD's 
perception system. Specifically, we aim to add adversarial 
points above the car, constrained physically, to make it 
undetectable by the fusion model.
Our evaluation metric is the attack success rate (ASR), 
which is the ratio of successful attacks against an object detector 
to the total number of attacks made. 
We also examine the relationship between recall and the IoU threshold 
during the attack. 
Moreover, we investigate the effects of the number, 
distance, and angle of injected adversarial points on the 
fusion model's performance.

\subsection{Experimental Setup}
We employ the KITTI~\cite{Geiger2013IJRR} dataset for simulation 
evaluation, a staple in training and testing 3D object detectors. 
As reiterated, our primary concern is car detection due to its 
critical safety implications in AD's perception system. 
The security of the LiDAR-camera fusion model MVX-Net~\cite{sindagi2019mvx} 
is evaluated using the implementation from MMDetection3D~\cite{mmdet3d2020}.

We set up the aforementioned object detectors in our lab, 
utilizing a server powered by an Intel Xeon Gold 6240C CPU @2.60 GHz, 
equipped with four GeForce RTX 3090 GPUs, and furnished with 
256 GB of RAM. This setup is utilized for optimizing adversarial 
point clouds.

\subsection{Number of Added Adversarial Points}
To discern the impact of different numbers of adversarial 
points on the fusion model's detection capabilities, we randomly 
select 100 cars from the KITTI dataset with the intent to render 
them undetectable. Fig. \ref{pointsnum} depicts the ASR's progression 
as a function of the number of adversarial points added. With 20 
adversarial points, the ASR stands at 39\%, 
but with 200 points, it escalates to 99\%. Generally, the ASR 
shows a positive correlation with the number of added adversarial points. 
However, its growth rate decelerates once the count surpasses 100. 
These findings corroborate that physically constrained adversarial 
points can effectively deceive the LiDAR-camera fusion model.
Furthermore, as can be seen from Figure 5, after the addition of 
adversarial points, the model's recall drops significantly. 
With the increase in the number of adversarial points, 
the recall is even lower at the same IoU. This indicates 
that the robustness of the MVX-Net~\cite{sindagi2019mvx} 
is notably compromised 
when interfered with by adversarial points.

\begin{figure}[t]
	\centering
	\includegraphics[width=0.42\textwidth]{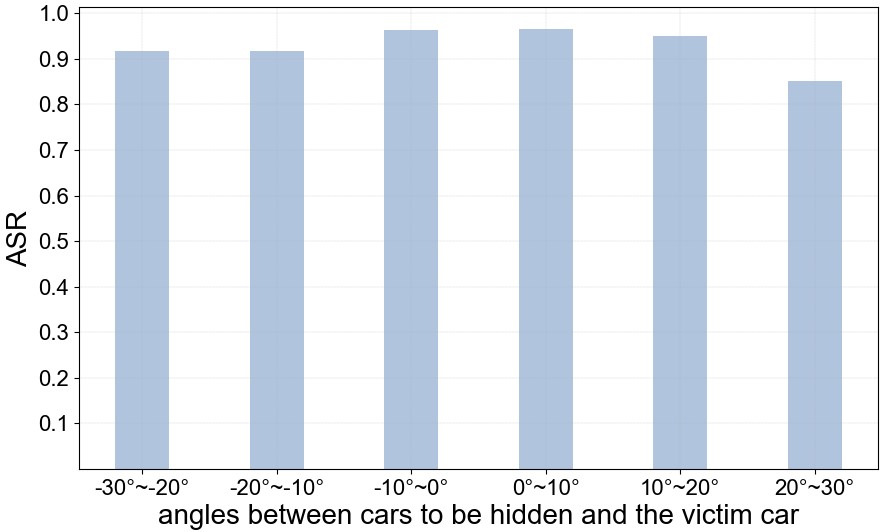}
	\caption{Attack success rate of spoofing fusion model
    with different angle between cars to be hidden and
    the victim car deployed with LiDAR.}
	\label{angle}
	\vspace{-5pt}
\end{figure} 

\subsection{Influence of Distance}
We explored the impact of distance between the cars intended to be 
concealed and the victim car equipped with LiDAR on the ASR. 
We randomly selected 1016 cars, which the fusion model detected with 
high confidence from the KITTI dataset, to serve as concealed targets. 
Among these, 263 cars are 5m-15m away from the LiDAR, 
376 cars are 15m-25m, 234 cars are 25m-35m, and 143 cars are 35m-45m. 
The results are shown in Fig. \ref{distance}.

Interestingly, the greater the distance of the adversarial 
points from the LiDAR, the higher the ASR. This suggests that cars 
farther away from the victim vehicle with LiDAR are easier to make 
undetectable by introducing adversarial points above them. This is 
because a closer car will typically send more reflected rays to the 
LiDAR, resulting in more recorded points and preserved features. 
Hence, the fusion model will detect the closer car with higher 
confidence, making it harder to conceal with adversarial points.

\subsection{Influence of Angle}
We also examined the impact of the angle between the cars intended 
to be hidden and the victim car equipped with LiDAR on the ASR. 
From the KITTI dataset, we randomly selected 950 cars detected 
with high confidence by the fusion model to be concealed.
Among these 950 cars, 
there are 73 cars to be hidden within the angle 
from -30° to -20° of the victim car,
158 cars within the angle from -20° to -10°,
282 cars within the angle from -10° to 0°,
204 cars within the angle from 0° to 10°,
139 cars within the angle from 10° to 20°,
94 cars within the angle from 20° to 30°.
The results are illustrated in Fig. \ref{angle}. The smaller 
the angle between the car intended to be concealed and the victim 
car (the more the car is directly in front of the victim car, 
with 0° being exactly in front), the higher the ASR. 
This aligns with adversarial expectations, as an attacker 
would ideally want the concealed vehicle directly in front of 
the victim vehicle, increasing the likelihood of abrupt braking 
and potential rear-end collisions.

\section{Conclusion}
In our research, we investigate the adversarial robustness of the 
LiDAR-camera fusion model in autonomous driving, particularly 
when injected with adversarial points that adhere to physical constraints. 
We discovered that by adding a limited number of adversarial points 
above a car, the fusion model can be deceived into not detecting the car. 
Intriguingly, this vulnerability exists even when only the LiDAR data 
channel is modified, without tampering with the image data channel. 
This finding highlights a significant safety concern in autonomous driving.
Our results indicate that the Attack Success Rate tends to 
increase with the quantity of introduced adversarial points. Moreover, 
the greater the distance between the adversarial points and the victim 
car equipped with LiDAR, the higher the ASR for concealing cars. 
Additionally, we observed that cars positioned closer to the front 
of the victim vehicle are more easily rendered undetectable by the 
fusion model when adversarial points are added.
We hope that our findings offer valuable insights into the robustness of 
multi-sensor systems, contributing to safer autonomous driving solutions.

\bibliographystyle{IEEEtranS}
\bibliography{IEEEabrv,refer}

\end{document}